\pdfoutput=1
\documentclass[sigconf]{acmart}

\usepackage{algorithm}
\usepackage{algorithmic}

\AtBeginDocument{%
  \providecommand\BibTeX{{%
    \normalfont B\kern-0.5em{\scshape i\kern-0.25em b}\kern-0.8em\TeX}}}

\setcopyright{acmcopyright}
\copyrightyear{2018}
\acmYear{2018}
\acmDOI{10.1145/1122445.1122456}

\acmConference[Woodstock '18]{Woodstock '18: ACM Symposium on Neural
  Gaze Detection}{June 03--05, 2018}{Woodstock, NY}
\acmBooktitle{Woodstock '18: ACM Symposium on Neural Gaze Detection,
  June 03--05, 2018, Woodstock, NY}
\acmPrice{15.00}
\acmISBN{978-1-4503-XXXX-X/18/06}

\begin{document}

\title{MMLN: Leveraging Domain Knowledge for Multimodal Diagnosis}




\author{Haodi Zhang}
\affiliation{%
  \institution{Shenzhen University}
  \department{College of Computer and Software Engineering}
  \city{Shenzhen}
  \country{China}}
\email{hdzhang@szu.edu.cn}

\author{Chenyu Xu}
\affiliation{%
  \institution{Shenzhen University}
  \department{College of Computer and Software Engineering}
	\institution{Second People's Hospital of Guangdong Province}
  \city{Shenzhen}
  \country{China}}
\email{2060271017@email.szu.edu.cn}

\author{Peirou Liang}
\affiliation{%
  \institution{Shenzhen University}
  \department{College of Computer and Software Engineering}
  \city{Shenzhen}
  \country{China}}
\email{2020155016@email.szu.edu.cn}

\author{Ke Duan}
\affiliation{%
  \institution{Shenzhen University}
  \department{College of Computer and Software Engineering}
  \city{Shenzhen}
  \country{China}}
\email{2019151087@email.szu.edu.cn}

\author{Hao Ren}
\affiliation{%
  \institution{Second People's Hospital of Guangdong Province}
  \city{Guangzhou}
  \country{China}}
\email{1810272037@email.szu.edu.cn}

\author{Weibin Cheng}
\affiliation{%
  \institution{Second People's Hospital of Guangdong Province}
  \city{Guangzhou}
  \country{China}}
\email{chwb817@gmail.com}

\author{Kaishun Wu}
\affiliation{%
  \institution{Shenzhen University}
  \department{College of Computer and Software Engineering}
  \city{Shenzhen}
  \country{China}}
\email{wu@szu.edu.cn}

%
%
%
%
%

\renewcommand{\shortauthors}{Trovato and Tobin, et al.}

\begin{abstract}
  Recent studies show that deep learning models achieve good performance on medical imaging tasks such as diagnosis prediction. Among the models, multimodality has been an emerging trend, integrating different forms of data such as chest X-ray (CXR) images and electronic medical records (EMRs). However, most existing methods incorporate them in a model-free manner, which lacks theoretical support and ignores the intrinsic relations between different data sources. To address this problem, we propose a knowledge-driven and data-driven framework for lung disease diagnosis. By incorporating domain knowledge, machine learning models can reduce the dependence on labeled data and improve interpretability. We formulate diagnosis rules according to authoritative clinical medicine guidelines and learn the weights of rules from text data. Finally, a multimodal fusion consisting of text and image data is designed to infer the marginal probability of lung disease. We conduct experiments on a real-world dataset collected from a hospital. The results show that the proposed method outperforms the state-of-the-art multimodal baselines in terms of accuracy and interpretability.
\end{abstract}




\maketitle

\section{Introduction}
%
%

\begin{figure}
  \centering
	\includegraphics[width=.5\textwidth]{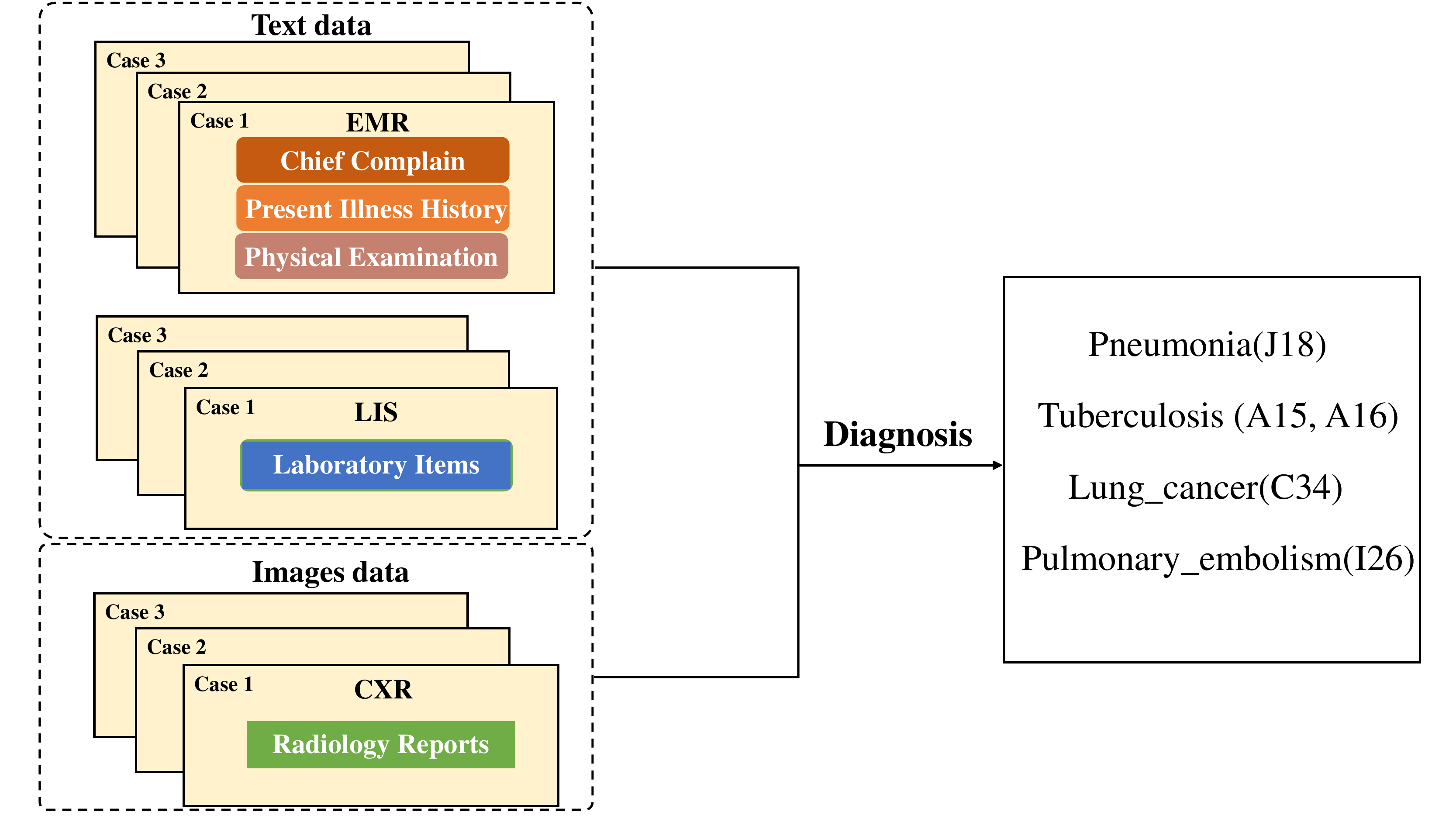}
  \caption{Disease diagnosis using multimodal data.}
	\label{fig:intro}
\end{figure}

In recent years, advances in deep learning and the release of multiple, large, publicly available chest X-ray (CXR) datasets have led to a promising performance in many medical imaging analysis tasks. Among all medical imaging approaches, CXR is the most common medical radiological examination in the world. As a particularly important modality, many lung diseases are first diagnosed by CXR. a variety of CXR applications have been researched and even boosted numbers of products.

However, CXR alone is not enough. Multimodality has been an emerging trend, integrating different forms of data such as text data and image data.(Figure~\ref{fig:intro}) 
Since the correlation between text modality and image modality is relatively large, diagnosis accuracy can be improved by these complementary data. 
Recently, success of transformer architecture in language domain has naturally led to its further application in cross-modal tasks involving language and vision. A transformers model used in multimodal settings, combining a text and an image to make predictions. The transformer model takes as inputs the tokenized text embedding and image embedding from a pretrained Resnet. These different inputs are concatenated to do classification. 

Besides making use of multimodal data, we leverage domain knowledge in machine learning model.  (Knowledge Representation and Knowledge Acquisition) It is necessary to fully integrate expert experience and domain knowledge into data-driven machine learning models in medical disease diagnosis. Domain knowledge can reduce the dependence of the learning model on labeled data, improve the interpretability of the model and enhance the model's robustness to uncertain data. Clinical guidelines are authoritative knowledge in the medical field. Experts are regularly organized to discuss and revise clinical guidelines, which are of great significance for disease diagnosis.

Incorporating domain knowledge, model interpretability can be improved. It is necessary to strengthen the interpretability of model prediction process and results for disease diagnosis task. Model interpretability is as important as prediction accuracy for medical diagnosis decision-making.

To leverage domain knowledge with multimodal data, we propose a Multimodal Markov Logic Network(MMLN) for lung disease diagnosis. In MMLN, we first formulate diagnosis rules in form of Markov logic according to authoritative clinical medical guidelines. Then, we learn each weight for every rules from numbers of text data such as EMRs, laboratory items and radiology reports. A pretrained model is finetuned to extract common lung pathology from CXR images. Finally, we infer the marginal probability of lung disease from the text and image data. 

We conduct experiments on a real-world dataset collected from the Second People's Hospital of Guangdong Province (GD2H) . The results show that MMLN can accurately diagnose lung diseases with good interpretability and outperform the state-of-the-art multimodal baselines. Our contributions are summarized as follow: 
\begin{itemize}
\item We formulate diagnosis rules in form of symbolic logic according to authoritative clinical medical guidelines, improving the accuracy and interpretability of the model.
\item We propose a data-driven and knowledge-driven framework called multimodal Markov logic network, which bridges domain knowledge and deep learning to help lung disease diagnosis.
\item We conduct experiments on a real-world dataset collected from a hospital to demonstrate the effectiveness of our model compared with the state-of-the-art baselines. 
\end{itemize}

\section{Related work}
The task of disease diagnosis has a long history. Since the last century, people have been exploring how to use computers to assist in disease diagnosis. MYCIN \cite{van1978mycin} is one of the first expert system that used rule-based reasoning to diagnose blood infections. Incorporating about 500 production rules and certainty factors, MYCIN performed at roughly the same level of experts. Besides, people start to research Clinical decision support system (CDSS). Kunhimangalam et al. proposed a CDSS for diagnosis of peripheral neuropathy using fuzzy logic. Through 24 input fields which include symptoms and diagnostic test outputs, they achieved 93\% accuracy compared to experts at identifying motor, sensory, mixed neuropathies, or normal cases. DXplain \cite{martinez2018diagnostic} is an electronic reference based CDSS that provides probable diagnosis based on clinical manifestations. In a randomized control trial involving 87 family medicine residents, those randomized to use the system showed significantly higher accuracy (84\% vs. 74\%) on a validated diagnosis test involving 30 clinical cases. The problem of pure symbolic models is the performance is limited, and the migration ability is relatively weak.
%


In recent years, deep learning has made a tremendous impact in the field of medical imaging, which can greatly enhance the capabilities of Image-level Prediction, Segmentation, Image Generation, Domain Adaptation, Localization, etc. With the release of several large publicly CXR datasets, including CheXpert \cite{irvin1901chexpert}, chestXray8 \cite{wang2017chestx}, PadChest \cite{bustos2020padchest}, etc., the development of CXR diagnosis is accelerating. A series of state-of-the-art CNNs for image-level classification are proposed. Among the pathology, pneumonia is one of the most studied subject. Singh et al \cite{singh2018deep} compare the performance of different architectures with various depths on a given task. Sirazitdinov et al \cite{sirazitdinov2019data} evaluate the effect of various data augmentation and input pre-processing methods. During the COVID-19, Wang et al. \cite{wang2021deep} proposed a deep learning pipeline for the diagnosis and discrimination of viral, non-viral and COVID-19 pneumonia from chest X-ray images with AUC of above 0.87. \cite{hou2021explainable} developed a explainable convolutional neural network based on CXR images classifcation and analysis for COVID-19 pneumonia detection. It can select instances of CXR images to explain the behavior to achieve higher prediction accuracy of above 96\%. Although pure machine learning based methods have a good performance, they only learn from data and output the probability. They all belong to probabilistic models, lack of the support of knowledge. 

There are some work incorporate knowledge graph (KG) or ontology as supplements to knowledge. For example, \cite{li2020real} propose a 8-step pipeline to build a medical knowledge graph (KG) with a novel quadruplet structure from EMRs. Then the KG is used to diagnosis recommendation, whose top 10 recall is 88.76\%. \cite{yin2019domain} propose a new model by directly introducing domain knowledge from the medical knowledge graph into an RNN architecture, as well as taking the irregular time intervals into account. Experimental results on heart failure risk prediction tasks show that the model not only outperforms state-of-the-art deep-learning based risk prediction models, but also associates individual medical events with heart failure onset. \cite{shen2018cbn} construct a clinical Bayesian network with complete causal relationship and probability distribution of ontology from EMRs data to enhance diagnostic inference capability. \cite{jiang2017learning} propose a Markov logic network with disease and symptom nodes from KG constructing from EMRs, which can learn and inference for medical diagnosis. Although knowledge is incorporated, it lacks theoretical basis and does not explore the intrinsic relationship between domain knowledge and multimodal data, but simply concatenate together. Compared with these methods, our model leverages the domain knowledge from authoritative clinical guideline, exploring the intrinsic relations between cross modality data, which differentiates it from the existing methods.

\section{Proposed Method}

\begin{figure*}
  \centering
	\includegraphics[height=.45\textheight]{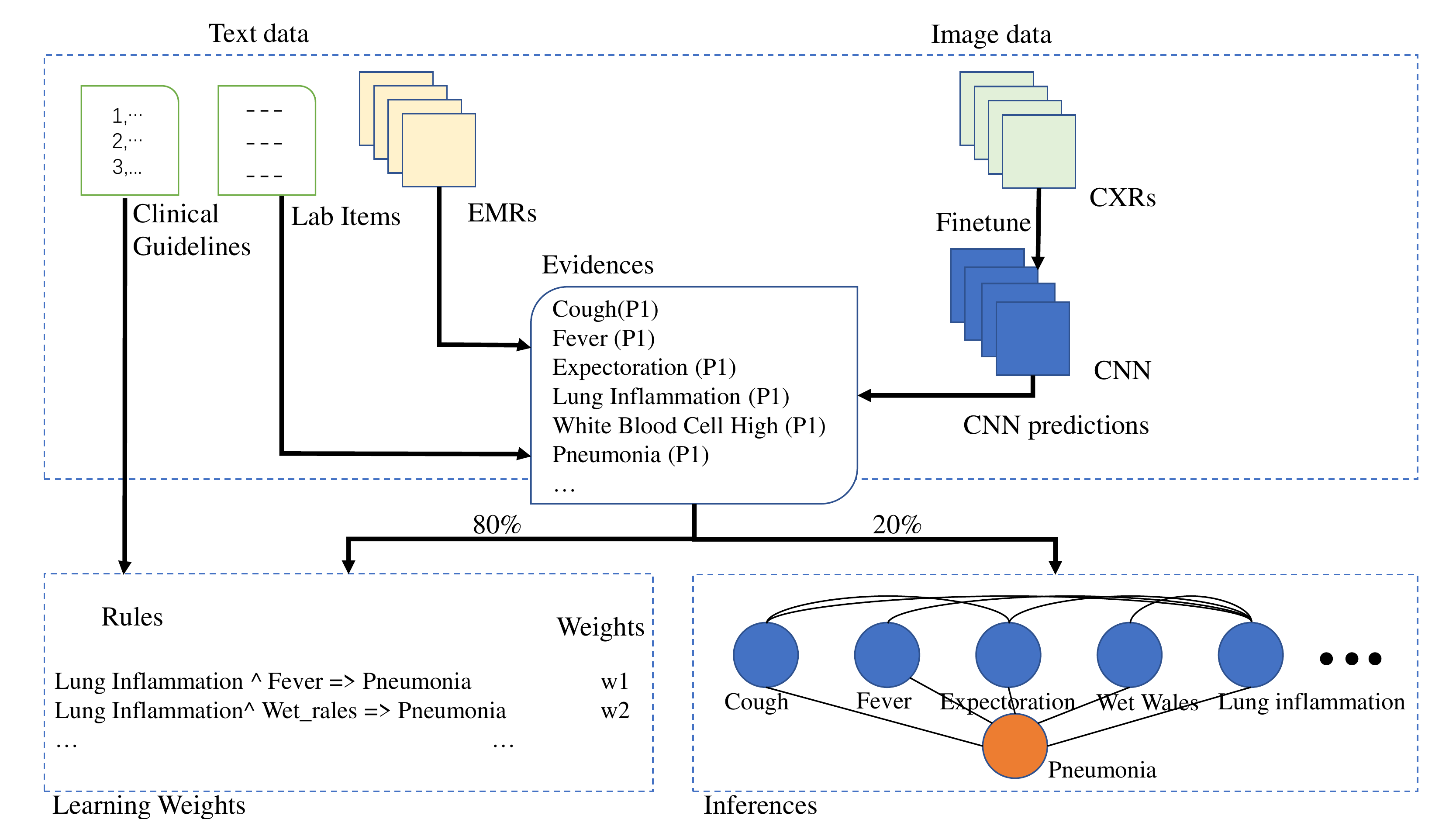}
  \caption{Multimodal Markov Logic Network framework.}
	\label{fig:framework}
\end{figure*}

We propose a simple yet effective multimodal Markov Logic Network that effectively integrates text data and image data (Figure~\ref{fig:framework}) . Text data mainly includes EMRs, laboratory items and radiology reports, while image data mainly refer to CXR images. Undoubtedly, multimodal data can improve the accuracy of disease diagnosis. Chest imaging examination is an indispensable part of doctor's diagnosis. Common lung diseases are first diagnosed using CXR. In this context, we effectively integrate multimodal data with the guide of clinical diagnostic criteria. 

\subsection{Formulate diagnostic rules}

\begin{figure*}
  \centering
	\includegraphics[width=\textwidth]{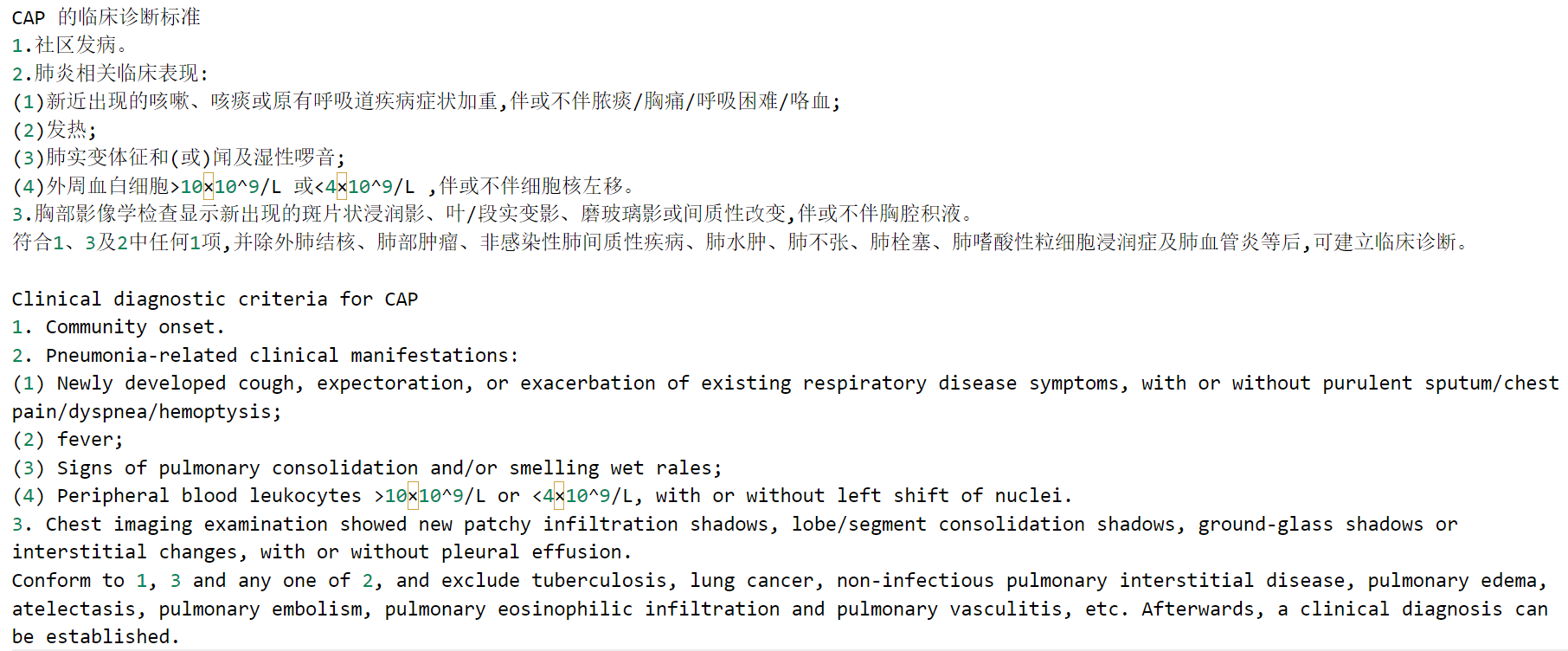}
  \caption{Clinical diagnostic criteria for CAP.}
	\label{fig:guildline}
\end{figure*}

Figure~\ref{fig:guildline} shows the clinical diagnostic criteria for community-acquired pneumonia (CAP). We use first-order logic (FOL) to represent knowledge about pneumonia clinical guidelines. 
For the first item, we define a query predicate called Pneumonia representing CAP. 
For the second item, we define evidence predicates for each clinical manifestation. For example, we define a predicate called Fever corresponding to 2 (2) . 
For the third item, we define evidence predicates about chest imaging examination. For example, we define predicates called Infiltration, Consolidation, etc.
We then use the defined predicates to construct the FOL formulas according to the logical relationship described in the last paragraph of the guideline. For the first item, all the selected pneumonia cases are CAP, so it is satisfied by default. For the second and third items, we define one of the rules as follow: 

$$ w \quad Fever (x) \land Consolidation (x) \implies Pneumonia (x) \quad (1) $$

However, the real world is complex and uncertain, especially in the domain of disease diagnosis. Therefore, we attach a weight w to each rule, which becomes a MLN. When a world violates one rule, it becomes less probable but not impossible. 

\subsection{Building multimodal evidence database}

\begin{algorithm}[tb]
	\caption{Building evidence database}
	\label{alg:evidence}
	\label{alg:}
	\textbf{Input}: $EMRs, CXR, Guideline$ \\
	\textbf{Output}: $Evidences$\\
	\begin{algorithmic}[1]
		\STATE \textbf{Begin} \textbf{\textit{formulate diagnostic rules}}\\
		\STATE Define the predicates from the $Guideline$ → $Predictates$\\
		\STATE Formulate rules using defined $Predicates$ according to the logic relations of $Guideline$ → $Rules$\\
		\STATE \textbf{End} \textbf{\textit{formulate diagnostic rules}}\\
		\STATE Finetune a $CNN$ model to extract pathologies from $CXRs$\\
		\STATE \textbf{Begin} \textbf{\textit{building evidence database}}\\
		\STATE Extract mentioned predicates about clinical manifestations from $EMRs$ → $Evidences\_EMRs$\\
		\STATE Extract mentioned predicates about chest imaging examination from $CXRs$  → $Evidences\_CXRs$\\
		\STATE Combine $Evidences\_EMRs$ and $Evidences\_CXRs$ to bulid $Evidences$\\
		\STATE \textbf{End} \textbf{\textit{building evidence database}}\\
	\end{algorithmic}
\end{algorithm}

\begin{figure*}
  \centering
	\includegraphics[height=.5\textwidth]{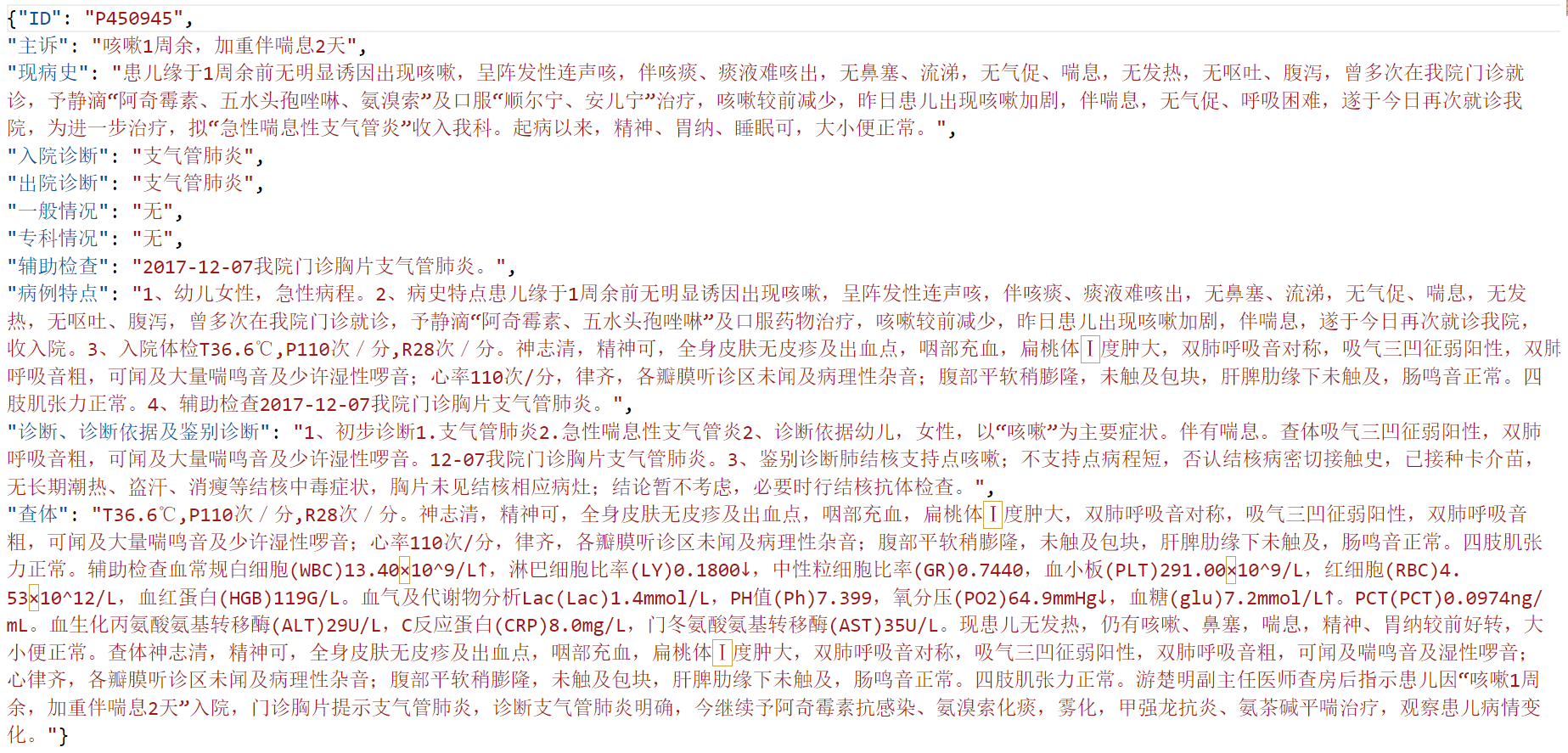}
  \caption{a sample of EMR.}
	\label{fig:EMR}
\end{figure*}

After formulating the diagnostic rules, our goal is to extract each possible grounding of each predicate appearing in rules to build evidence database for weight learning and inference. The detailed process is shown in Algorithm \ref{alg:evidence}.

\subsubsection{Text data}
	We extract clinical manifestations from text data. After the regular expression text matching, we convert the EMRs from unstructured text to semi-structured Json format with some important fields. Figure~\ref{fig:EMR} is an example of Json format EMRs, from where we extract grounding predicates. 
We first collect a dictionary of common symptoms of lung diseases. Then a modified version of NegEx \cite{chapman2001simple} and NegBio \cite{peng2018negbio} was used to extract positive symptoms.
Specifically, we extract positive symptoms from chief complain and present illness history, physical signs from physical examination, laboratory items from structured laboratory information system (LIS) and so on. 
For example, we extract the following grounding predicates from Figure~\ref{fig:EMR}: 

White\_blood\_cells\_high(P450945)

Wet\_rales(P450945)

Cough(P450945)

Expectoration(P450945)

Pneumonia(P450945)

P450945 refers to patient ID. 

\subsubsection{Image data}

\begin{figure}
  \centering
	\includegraphics[width=.4\textwidth]{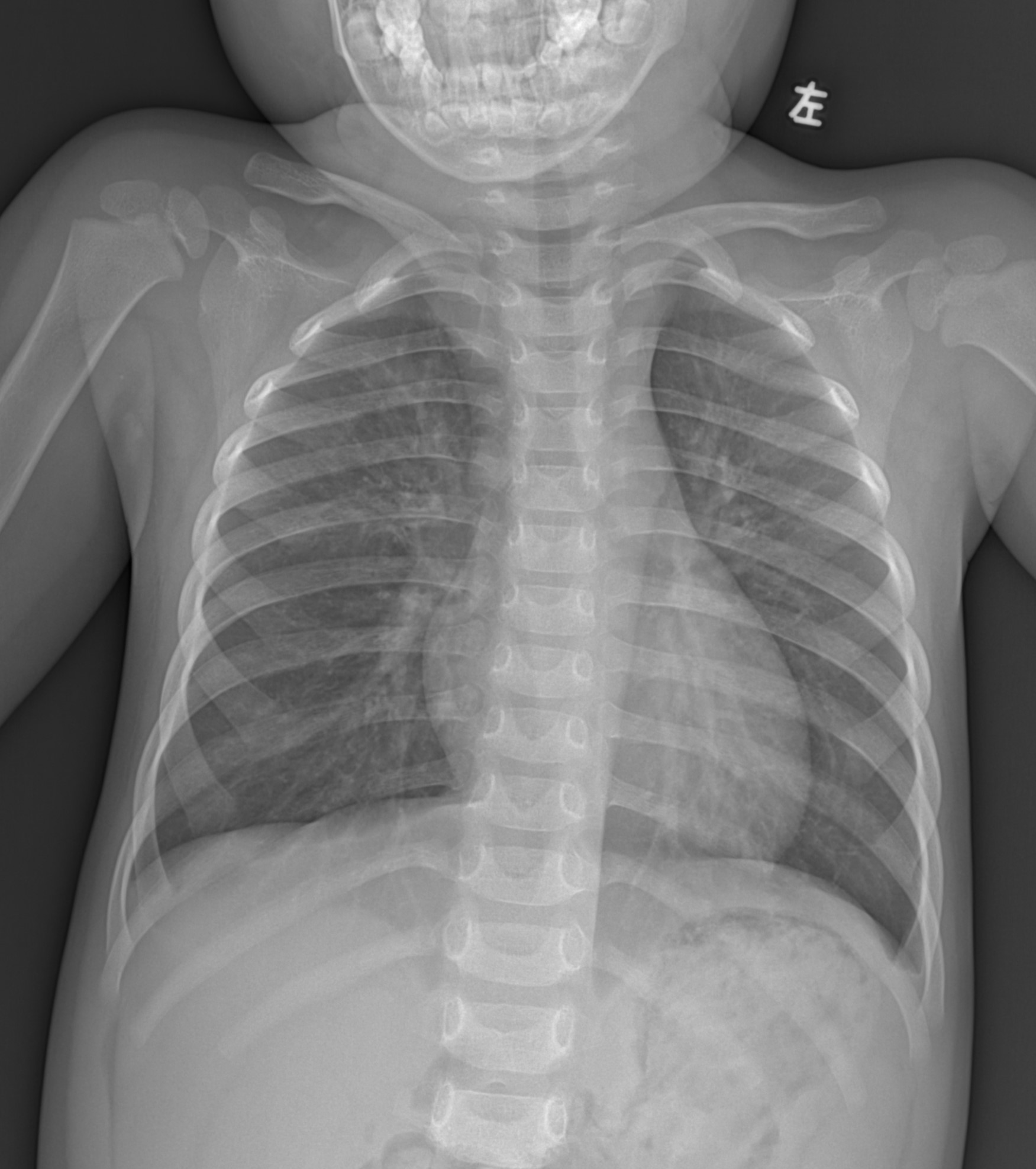}
  \caption{a sample of CXR image.}
	\label{fig:CXR}
\end{figure}

We extract chest imaging examination from image data. We use a pretrained multi-label classification CNN model, which has been trained on several large publicly available CXR datasets. The model takes CXR images as input and output the probability of common lung pathologies such as Infiltration, Consolidation, etc.
	We collected the CXR images of the hospital and corresponding labels to finetune the model. A modified version of Chexpert-labeler \cite{irvin2019chexpert} was used to extract pathologies as the ground truth labels from radiology reports.
	After finetuning, we regard the CNN predictions as grounding predicates about chest imaging examination. Since CNN prediction is a probabilistic value in [0, 1], we set a threshold of 0.6 to extract the positive grounding predicates. 
For example, we pass Figure~\ref{fig:CXR} through finetuned models and output positive pathology: CXR\_Lung\_inflammation. Therefore, we get a grounding predicate as follows:
 
CXR\_Lung\_inflammation(P450945). 

Combining the grounding predicates extracted from the text data and image data, we get an evidence database, which is a large scale of observations from real-world multimodal data. 

\subsection{Learning weights and inference}

\begin{algorithm}[tb]
	\caption{Weights learning and inference of MMLN}
	\label{alg:learn_infer}
	\textbf{Input}: \textit{Predicates, Rules, Evidences} \\
	\textbf{Output}: \textit{Pneumonia\_probability}\\
	\begin{algorithmic}[1]
		\STATE Initialize Nodes for each possible grounding of each predicate in $Predicates$\\
		\STATE Add one feature for each possible grounding of each rule in \textit{Rules}\\
		\STATE Split the \textit{Evidences} into training set \textit{Evidences\_train} and test set \textit{Evidences\_test}\\
		\STATE \textbf{Begin} \textbf{\textit{learning weights}}
		\STATE Initialize the weights of \textit{Rules} by zero
		\STATE $Learnwts(Predicates, Rules, Evidences\_train)$ → $Weights$
		\STATE \textbf{End} \textbf{\textit{learning weights}}
		\STATE \textbf{Begin} \textbf{\textit{inference}}
		\STATE  \textbf{for} $case_i$ in Evidences\_test \textbf{do}
		\STATE $Infer (Predicates, Rules, Weights, Evidences\_t$ → $Pneumonia\_probability_i$\\
		\STATE  \textbf{end for}
		\STATE \textbf{End} \textbf{\textit{inference}}
	\end{algorithmic}
\end{algorithm}

After building the evidence database, we split the the database into training set and test set in a 80/20 ratio. Training set is used to learn weights, while test set is used to inference. The detailed algorithm is shown in Algorithm~\ref{alg:learn_infer}. 

\subsubsection{Learning weights}
We use the MLN tool Alchemy \cite{kok2005alchemy} to learn each weight of rules and inference. 
Given defined predicates, unweighted rules and the training set evidence database, we use function learnwts to learn each weight of rules. For example, we learn each weight of rule as follows: 

\begin{figure*}
$$ 0.933901 \quad Lung\_inflammation(x)  \land Cough(x) \land Expectoration(x) \implies Pneumonia(x) $$

$$ 3.59945 \quad Lung\_inflammation(x)  \land Fever(x) \implies Pneumonia(x) $$

$$ 3.49846 \quad Lung\_inflammation(x)  \land Wet\_rales(x) \implies Pneumonia(x) $$

$$ 1.55942 \quad Lung\_inflammation(x)  \land White\_blood\_cells\_high(x) \implies  Pneumonia(x) $$

$$ 1.24346 \quad Lung\_inflammation(x)  \land White\_blood\_cells\_low(x) \implies  Pneumonia(x) $$
\end{figure*}


Higher weight represents more probable rule.

\subsubsection{Inference}
The goal of inference is to calculate the marginal probability of specific illness to make a disease diagnosis. 
Given defined predicates, weighted rules and the test set of evidence database, we use function infer to perform inference. As for the example mentioned in section 3.2, we get the result as follows: 

Pneumonia(P450945) 0.99895, 

the probability represents the illness probability, it shows that the case is highly pneumonia.

\section{Experiment}

\subsection{Dataset}
We collected the multimodal data of the admitted patients from the EMR system and Picture Archiving and Communication System (PACS) of the hospital information department. The text data consists of EMRs, laboratory items and radiology reports. The image data mainly includes CXR images. The de-identified data collected in this experiment all come from real medical data from the Second People's Hospital of Guangdong Province, and the privacy of patients has been removed.

\subsubsection{Data processing}
We first find out all cases with both EMR and CXR. In our experiment, disease diagnosis is regard as a binary classification task of pneumonia and other diseases related to pneumonia. We treat pneumonia cases as positive samples, and other lung diseases as negative samples. 
Since doctors usually give a series of diagnostic codes rather than a single disease, we need to screen the cases. For pneumonia cases, we select cases diagnosed with pneumonia in admission diagnosis to ensure that they were CAP cases. We then further screened the cases with the main diagnosis of pneumonia from the discharge diagnosis and preliminary diagnosis. For negative samples, it is also necessary to additionally exclude cases with pneumonia in the diagnosis to ensure that they does not suffer from pneumonia. 

\subsection{Comparison of unimodal model and multimodal model}
In this section, we compare the performance between unimodal and multimodal models. 
Unimodal means that we build evidence database purely on text data. Then we use the unimodal database to learn weights and inference.

As for the clinical manifestations, we adopt the same methods mentioned in section 3.2 to extract grounding predicates. As for the chest imaging examination, we extract grounding predicate Lung\_inflammation from EMRs instead of CXR images. We use regular expressions to match related description from EMRs‘ fields such as physical examination, auxiliary examination, case characteristics, diagnosis, diagnostic basis and differential diagnosis, general situation, specialty situation. 
Multimodal model is implemented in multimodal disease diagnosis framework to effectively incorporate text data and image data. A multi-label classification pretrained model on TorchXrayVision \cite{Cohen2020xrv} is used to output common lung pathologies. 

\begin{figure}
  \centering
	\includegraphics[width=.5\textwidth]{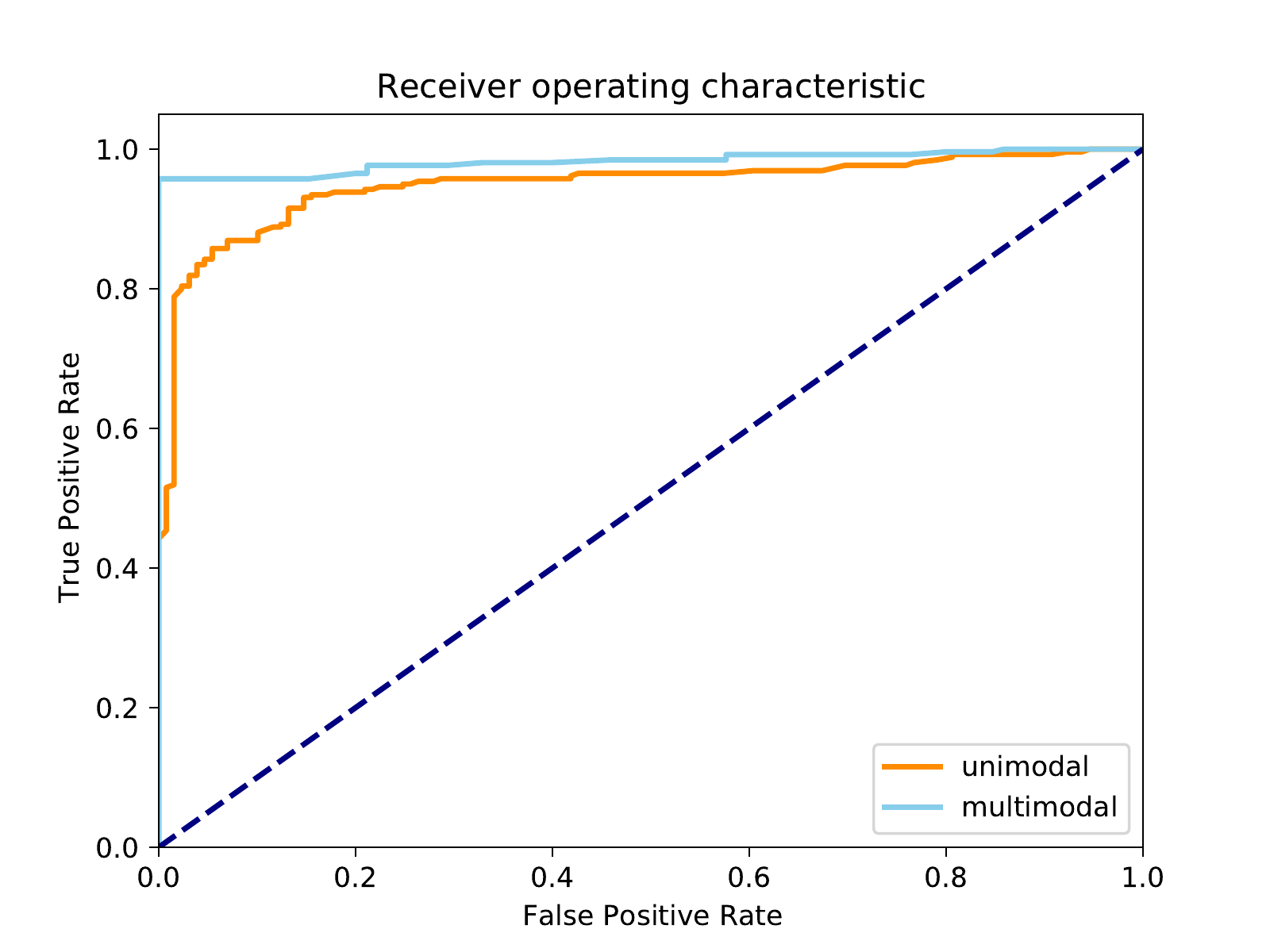}
  \caption{ROC of unimodal model and multimodal model.}
	\label{fig:uni_multi}
\end{figure}

After learning weights and inference, we get the unimodal and multimodal inference accuracy respectively. The performance is shown in Table~\ref{tab:performance} and ROC is compared in Figure~\ref{fig:uni_multi}. 
As shown in Table~\ref{tab:performance}, the overall performances of multimodal model are better than those of the unimodal model. In particular, our unimodal model reaches an accuracy above 0.90. Incorporating image data, our multimodal model has a large improvement in accuracy, which is near to 0.97. 
We also draw the ROC curves for the classification results of unimodal and multimodal model, as shown in Figure~\ref{fig:uni_multi}. We can see that the curve of multimodal is closest to the left top corner, and its corresponding AUC score is 0.965 which is also the highest AUC values. 
Generally, the results prove that multimodal data can actually improve the accuracy of disease diagnosis. 

\begin{table*}[ht]
	\centering
	\caption{Performance of our model against competitive methods}
	\label{tab:performance}
	\begin{tabular}{lrrrrr}
		\toprule
		Model  &Accuracy & AUC & F1 &  Precision & Recall \\
		\midrule
		MMBT(Bert-base-uncased) & 0.897 & 0.883  & 0.916 & 0.873 & 0.963 \\
		MMBT(Bert-large-uncased)  & 0.950 & 0.946 & 0.958 & 0.946 & 0.970 \\
		ConcatBert  & 0.936 & 0.929& 0.946 & 0.924 & 0.970 \\
		Ours(Unimodal) & 0.905 & 0.948 & 0.929 & 0.927 & 0.931 \\
		Ours(Multimodal) & 0.965 & 0.983 & 0.976 & 1.000 & 0.954   \\
		Ours(Multimodal, small size) & 0.924 & 0.967 & 0.948 & 0.987 & 0.911   \\
		\bottomrule
	\end{tabular}
\end{table*}

\subsection{Comparison of MMLN and multimodal baselines}
We compare three multimodal baselines with MMLN in this experiment. 
MultiModal BiTransformers (MMBT) \cite{kiela2019supervised}: the underlying pretrained models are BERT \cite{devlin2018bert} and ResNet \cite{He2015}. The text data is passed through BERT to generate word embedding, while the image data is passed through ResNet to generate CNN embedding, and then the two embeddings are concatenated and passed through a classifier for disease classification. We use two models with different size to compare.
ConcatBert \cite{arevalo2017gated}: it is another model for multimodal classification with text and images, whose text representation obtained from pretrained BERT base model and image representation obtained from VGG16 pretrained model.

For the text data, we extracted the main complaint, current medical history, physical signs, laboratory items, lung pathologies from EMRs. We concatenate these fields of text and translate it into English as text data input. For the image data, since multiple CXR images are generated as one patient has done CXR examinations by multiple times or multiple angles, we just randomly select one of the CXR images as the image data input. We select cases with both EMR and CXR, including a total of about 2000 cases. The performance is shown in Table~\ref{tab:performance}. 

As shown in Table~\ref{tab:performance}, the accuracy and F1 score of our multimodal model are better than those of the baseline models. Specifically, three classical deep learning models achieve accuracies in the range of 0.89 to 0.95. At the same time, except MMBT with bert-base-uncased, the baselines also yield accuracies with values from 0.93 to 0.95. In comparison, our multimodal reaches an accuracy near to 0.97. The same trend can also be observed in the comparison results of F1 scores. Generally, the performance of our multimodal model is superior to those of the other competing methods based on deep learning.

\subsection{MMLN robustness analysis}
When we collect data in the hospital, we find that it is quite difficult to obtain multimodal data as there is a widespread problem of missing modalities. Therefore, we explore the effect of training set size on MMLN here to illustrate the model is robust on dataset size.
 
Our original size of cases, including pneumonia cases and other pneumonia differential diagnosis cases, is about 2400, which is a medium-sized dataset. In this experiment, we only take 1/10 of the number of each disease case, and performed weight learning and inference on a small size of dataset.

\begin{figure}
  \centering
	\includegraphics[width=.5\textwidth]{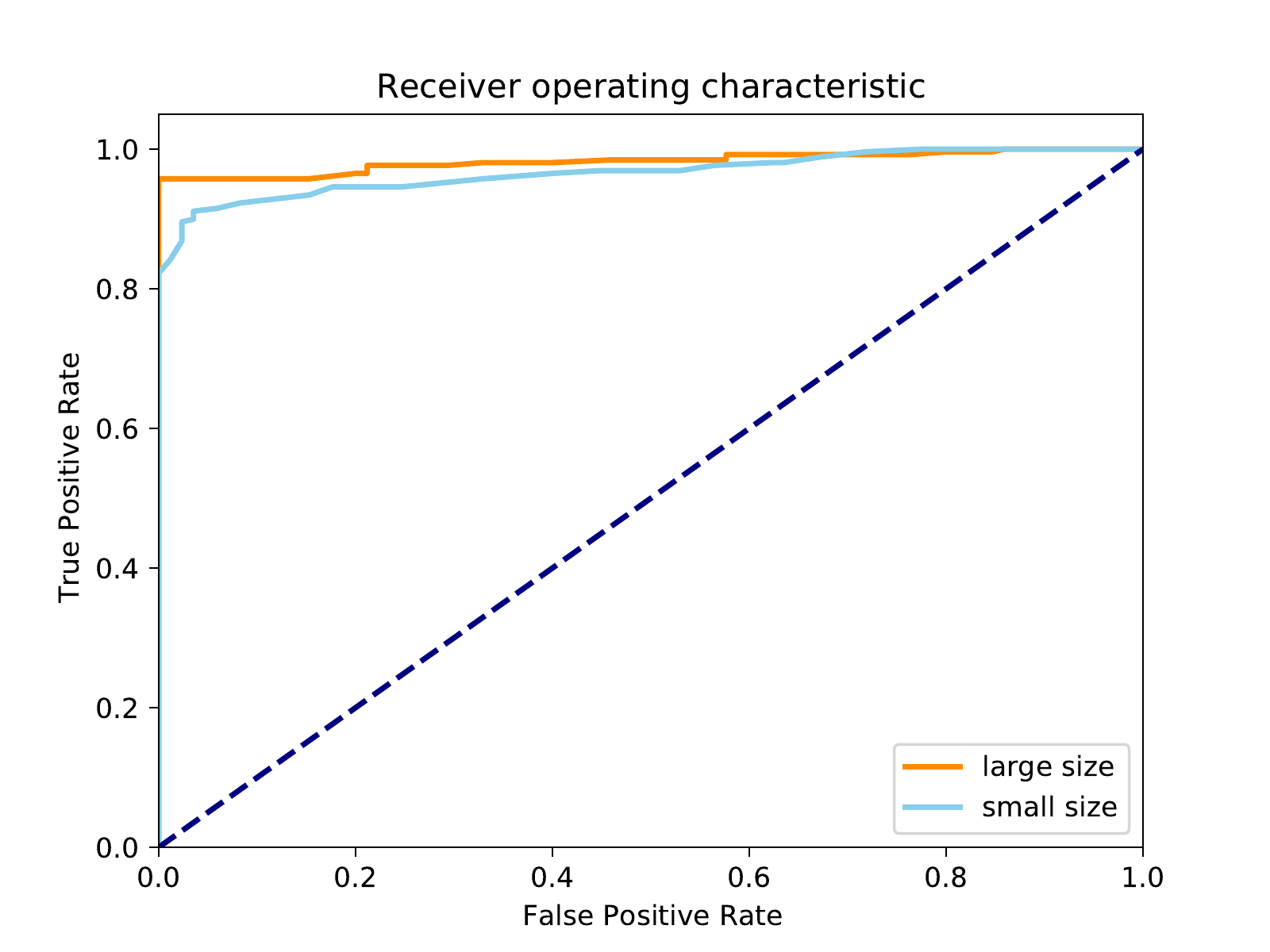}
  \caption{ROC of multimodal model learning on large and small size dataset.}
	\label{fig:large_small}
\end{figure}

In Table~\ref{tab:performance}, we can see that weights learning by large size dataset are bigger than those in small size. 
Table~\ref{tab:performance} lists the results of large size and small size multimodal model performance. Even if the size is only 1/10 of the original, the performance will not drop too much, the accuracy reaches 0.924, and the F1 score reaches 0.948, which is even better than MMBT (bert-base-uncased). In Figure~\ref{fig:large_small}, two ROC curves are very close. In general, MMLN has little effect on the size of the training data, and even if the size is small, we can get a not too bad performance.

\section{Conclusion}
In this paper, we propose a data-driven and knowledge-driven MMLN model for lung disease diagnosis. We formulate diagnosis rules in form of Markov logic according to authoritative clinical medical guidelines, which improves the accuracy and interpretability of the model. Moreover, MMLN effectively fuse multimodal data and reveal intrinsic relations between different data sources by leveraging the medical domain knowledge. 
The experimental results show that knowledge and data is fully complementary to better the downstream diagnosis task and MMLN outperforms the state-of-the-art multimodal models for disease classification.

\section{Acknowledgments}
\begin{acks}

\end{acks}



\end{document}